Accepted for Publication in **Philosophy of Translation**

# Translation Indeterminacy and the Distributional Fallacy

Michael Carl
mcarl6@kent.edu
Kent State University
Modern and Classical Language Studies

## Abstract

Large language models (LLMs) are commonly associated with the distributional hypothesis, according to which (1) semantic meaning is grounded in distributional patterns of linguistic context, and (2) knowledge of cross-linguistic distributional correspondences allows for successful translation. This paper rejects the first claim as a causal inversion: linguistic distributions reflect patterns arising from meaning-making practices rather than constituting their source. At the same time, it accepts the second claim, arguing that translation—human or machine—can succeed without requiring access to meaning or reference. Knowledge of interlingual distributional correspondence and their inferential organization may be sufficient for translation. The paper develops an ecological-enactivist perspective, according to which reference and meaning are grounded in agent–environment interaction and stabilized through action-grounded concepts—forms of world-involving cognition that current LLMs do not possess.

## 1 Introduction

Harris (1954) suggested what is known as the *distributional hypothesis* (DH):

> If two expressions are semantically similar, they tend to occur in similar collocational distributions.

which, in logical terms, corresponds to:

Semantic similarity ⇒ Distributional similarity

However, Harris maintained that "Meaning is not a unique property of language, but a general characteristic of human activity." and that "the structure of language does not necessarily conform to the structure of … meanings" (ibid 151). While the distributional structure of language "correlates in some way with the substance of what is being said," it does not conform "in some one-to-one way with some independently discoverable structure of meaning" (152) i.e., "there is no independently-known structure of meanings which exactly parallels linguistic structure." Harris also discussed a reversed, abductive form of his DH as a *discovery procedure*: distributional similarity can be used as a diagnostic or proxy for meaning similarity but was never intended as a constitutive theory of meaning:

Distributional similarity ⇒ Semantic similarity

Firth (1957) builds on this abductive DH version stating, "You shall know a word by the company it keeps", which has then sometimes been taken not merely as a methodological tool for the analysis of semantic similarity, but as in fact for constituting meaning. As an empirical or abductive tool, the reversed DH can be highly effective for predicting linguistic organization, such as in large language models (LLMs). The conceptual problem arises when this predictive relationship is elevated into a constitutive theory of meaning. It becomes a logical fallacy, when as Sahlgren (2005) posits "distributional representations do constitute full-blown accounts of linguistic meaning", i.e., when what is useful for detecting or predicting linguistic organization is taken to be what constitutes meaning. I call this the distributional fallacy (DF). I argue that distributional patterns can capture linguistic regularities without necessarily accessing the relational and inferential organization of meaning itself. The DF conflates an epistemological relation (how semantic similarity can be detected) with a metaphysical one (what grounds meaning).

Variations of the DF are evident in numerous formulations of contemporary NLP. Lenci (2018), for instance, writes that "the semantic similarity of lexical items is a function of their distribution in linguistic contexts," or Evert (2008) makes out “that distributional similarity is semantic similarity, or at least that the former can be used to construct a representation of the latter."

Curiously, the DF stands in direct opposition to Quine's (1960) indeterminacy thesis. Quine's (1960) thesis concerns the impossibility of uniquely determining the semantic or referential content underlying linguistic behavior. There is no empirically accessible “fact of the matter”, he says, that uniquely fixes what a speaker means or what a linguistic expression refers to: distributional or behavioral similarity does not entail uniquely determined semantic similarity:

$$\text{Distributional similarity} \nRightarrow \text{Semantic similarity}$$

If the available empirical evidence does not uniquely determine semantic content and if translation is understood as the recovery and preservation of that content in another language, then we cannot uniquely determine which target language expression preserves the source content. Thus, for Quine, the indeterminacy of meaning entails the indeterminacy of translation: “there can be no determinate and uniquely correct translation, meaning, and reference for any linguistic expression.”[1]

Grounding and reference of linguistic expressions have been a central topic in the theory of mind for decades, as they are believed precursors for successful communication, translation and truth. Underlying by far most of these approaches is the idea that language is fundamentally representational, about standing for objects in the world, and for LLMs to be successful, they would need to be representational, too, in a related sense. I will address the idea that LLMs might fix reference through descriptive designators (Searle 1958) as well as with Kripke’s (1979) notion of rigid designation and causal communication chain.

I will then formulate an ecological-enactivist view which maintains that meaning emerges through embodied agent-environment interaction and that this interaction is sedimented in bodies of language. Language, in this view, is primarily a means of controlling and coordinating action, directing attention, participating in practices, and opening affordances in which reference becomes secondary. That is, the ecological-enactive account suggests that reference is a historically stabilized achievement of embodied coordination of human communities, where language exhibits statistical regularities because people have repeatedly interacted with the world in similar ways. But before speakers can successfully refer, they must already participate in practices that make names, descriptions, and communicative coordination possible. These practices are structured by affordances and, in linguistic communities, by enlanguaged affordances (Kiverstein & Rietveld 2018).

From this ecological perspective, Kripke's causal communication chain can be reinterpreted as the historical stabilization and transmission of shared affordance landscapes, rather than merely as the transmission of names, while Searle's sets of descriptive designators can be understood as resources that speakers exploit in situated communication. Once communicative practices become sufficiently stabilized, they leave behind linguistic traces that encode aspects of their inferential organization. From these traces, LLMs can learn, exploit, and reproduce the inferential organization, discourse regularities, and communicative conventions without participating in the embodied interactions and relevance realizations through which these practices acquired their meaning. The success and widespread use of LLMs suggest that much of what is required for communication and translation can be achieved by exploiting those historically stabilized collocational and inferential regularities, without reconstructing the determinate referential content that originally gave rise to them. The DH can therefore be reformulated into a more defensible prediction hypothesis in which distributional regularities provide access to the predictive linguistic organization required for successful translation.:

$$\text{Distributional regularity} \Rightarrow \text{Predictive linguistic organization}$$

While this does not resolve Quine's indeterminacy thesis it weakens its implications for translation: translation need not uniquely determine reference in order to reproduce functionally appropriate linguistic behavior. This also resonates with observations from some translation scholars who have noted, for a long time, that the distributional profile of language usage may be sufficient for translation.

## 2 The Sedimentation of Meaning in Language

LLMs acquire extraordinarily rich, high-dimensional regularities from language, yet the objective that drives this learning remains fundamentally distributional. During training, models learn to predict the next (or a masked) token from its linguistic context. They are therefore exposed to increasingly sophisticated patterns of collocation, syntax, discourse, and inference, but not to the perceptual, embodied, or social interactions through which human concepts are originally acquired. A pure LLM never encounters the objects, events, or practices that language is about.

[1] https://iep.utm.edu/indeterm/

Thus, what is often described as "representation of concepts" in LLMs may be more accurately understood as a highly compressed model of the collocational and inferential structure of language. These structures resemble conceptual organization because human language itself is the product of generations of embodied, socially situated agents whose interactions with the world have left systematic regularities in linguistic practice. Rather than recovering concepts directly, LLMs recover the statistical imprint that human communication leaves in text. An LLM model shadows that meaning casts onto language rather than meaning itself.

Searle's (1980) Chinese Room argument and Harnad's (1990) Symbol Grounding Problem identify a similar common limitation: symbols that relate only to other symbols do not thereby acquire semantic content. Considered in isolation, the linguistic world of an LLM forms a self-contained network in which tokens acquire their significance solely through their relations to other tokens. There is no direct coupling to the referents, affordances, or embodied practices from which those relations originally emerged.

Human conceptual competence, by contrast, develops through multiple, mutually reinforcing forms of grounding. It is shaped by perceptual engagement with the environment, by participation in shared social practices that stabilize reference and meaning, and by the capacity to coordinate attention and action with others. Language itself is therefore not the source of meaning but the historical product of embodied relevance realization and communicative interaction. From this perspective, the inferential organization captured by LLMs is best understood as the culturally sedimented outcome of human meaning-making rather than its origin. This distinction is central to the argument developed below: while distributional structure may be insufficient to explain the emergence of meaning, it may nevertheless prove sufficient to support successful translation by preserving the inferential organization that embodied linguistic communities have collectively stabilized over time.

## 2.1 Rigid Designation

Kripke and Putnam established semantic externalism: "meaning ain't in the head." (Putnam 1975). In *Naming and Necessity*, Kripke (1979) argues that the reference of proper names and natural kind terms like "water" or "gold" is fixed not by descriptive content but by a causal-historical chain linking current use back to an original dubbing event where the term was introduced in the presence of its referent. When I use the name "Aristotle," according to Kripke, I refer to a particular historical person not because I possess any uniquely identifying description, but because there is an unbroken causal chain tracing back to people who were in actual contact with him. Something similar applies to natural kind terms: "water" refers to $H_2O$ not because speakers know the chemical formula, but because there is a causal chain connecting uses of the word back to actual samples of the substance.

LLMs are causally downstream of the linguistic behavior of reference-capable agents: the collocational profile of "Aristotle" was shaped by texts written by people embedded in the causal-historical chain that fixes reference to the ancient philosopher. In this minimal sense, an LLM's token of "Aristotle" is causally connected to Aristotle, no less than any ordinary speaker's use of a name they acquired only through testimony (cf. Mollo and Millière 2026) & . Whether this makes the LLM itself a link in the causal communicative chain, rather than a compression of many other people's links, is contested: on one reading the LLM's reference, if it has any, is parasitic on the reference already secured by its training data ("derivative reference"), rather than a new instance of a competent human speaker. I take no stand here on whether that further step is available to LLMs, but the substantive disagreement lies elsewhere.

It is, I think, in whether the LLM's use of "Aristotle" is rigid rather than descriptive, and this is independent of the causal-chain question. A rigid designator picks out the same individual in every counterfactual scenario in which that individual exists; a definite description (e.g., "the teacher of Alexander") does not, since it would pick out someone else in a scenario where Plato's other student tutored Alexander. So the relevant test is not "does the token have causal ancestry"[2] but: does the model's use of the name track the individual across counterfactual variation in the associated descriptions, or does it simply track whichever entity currently best satisfies the cluster? On the latter behavior — treating "Aristotle" as interchangeable with "whoever taught Alexander and studied under Plato," such that the name would shift reference if the model were told these properties belonged to someone else — the LLM would be behaving exactly as Kripke's target descriptivist does and would be failing to secure rigid reference regardless of whether the underlying tokens are causally connected to Aristotle. I will argue for the latter.

[2] Baggio and Elliot (2026) argue that with subword tokenization, "there is nothing natural nor historical about strings of symbols" in LLMs.

## 2.2 Descriptive Designation

Some authors deny that LLM tokens are connected to their referents by a Kripkean communication-chain at all (Baggio & Elliot 2026). Wheeler (2025) points out that LLMs do not process their training data as sentences with intended referents: LLMs extract word-sequence probabilities from texts the way a corpus linguist might count occurrences of "and," indifferent to whether the text expresses a proposition. On this view there is no chain here to be preserved, statistically or otherwise, it is simply absent. What secures meaning and truth-value for LLM outputs is, for Wheeler, *design-intention* rather than causal-historical continuity: in Millikan's (1984) terms, an LLM has a derived proper function assigned by its designers; in Searle's (1980) terms, its outputs are derived from the intentionality of the people and institutions that design, deploy, and interpret the system.

LLMs do not “remember posts” but encode knowledge in huge probabilistic networks. LLMs compress all human text into a single predictive engine, Wheeler says. In order to do so they must compress the text into a probabilistic latent structure that encodes grammar, logic, causality, narrative flow, etc. During inference, each token prediction is conditioned by this latent structure. According to an LLM self-report, quoted in Wheeler (2025), LLMs produce “an immensely detailed representation of meaning and context synthesized from the entire conversation so far” and answers are not ”pulled from storage, but generated in real-time based on patterns learned during training.”

Cut off from any connection to the intentional acts of the humans who produced the training texts, an LLM's use of a name reduces, in Wheeler’s terms, to something close to Searle's (1958) cluster-of-descriptions theory, the very view Kripke's causal communication-chain was designed to displace. That is, LLMs end up treating names as predicates, i.e., bundles of associated descriptions, rather than rigid designators.

If this is right, LLMs may be described as systems whose tokens are causally connected to their referents via the causal history of the training corpus, but not via any chain of referring intentions. The use of LLM tokens remains descriptive rather than rigid, securing, in Frege's terms (1892; trans. Black, 1948), only the sense-cluster, and only contingently and derivatively, via design-intention, the reference that cluster happens to track.

## 2.3 Action-grounded Concepts

Humans often use action-based concepts they cannot fully articulate. A child knows how to ride a bike long before they can explain balance and momentum. A radiologist can identify a tumor before they can verbalize all the visual cues they're using. There is a kind of knowledge — perhaps what Ryle (1949) called "knowing how" or what Polanyi (1967) called "tacit knowledge" — that is real and causally efficacious but not propositional.

Squirrels, for instance, have this kind of knowledge in abundance. A squirrel has a concept of what it means to jump from one branch to another, to search for nuts, to chase a conspecific mate etc. which, however, it cannot express in language. The squirrel's action-grounded concept of "branch" is not linguistic or propositional, it is sensorimotor. The squirrel knows branches as graspable surfaces, launch points for jumps, structures with a particular flexibility and weight-bearing properties. This knowledge is manifested in skilled behavior: the squirrel adjusts its jump trajectory based on branch thickness, wind conditions, distance to target, etc.

The squirrel has practical knowledge, Ryle’s know how, that is causally connected to the world through perception and action. This knowledge is conceptual as it allows for 1) predictive modeling 2) generalization, 3) misrepresentation and 4) graded response. If one changed the physics of branches (made them all slippery, or elastic like rubber), the squirrel's action concept would be falsified by experience. The squirrel would fail to land jumps, get confused, and potentially update its behavior. The action concept has consequences that can be wrong.

## 2.4 Inferential Concepts

LLMs have something like the inverse: they can articulate (generate linguistically coherent text about) linguistic concepts of which they have no practical grasp, whatsoever. They have the linguistic form of knowledge without substance. An LLM's concept of a branch is a purely collocational profile: “branch” patterns with tree, leaf, climb, fall, wood, bird, nest, snap, etc. These patterns are learned from text produced by humans who do have action-grounded concepts of branches.

LLMs can perform valid inferences within the collocational structures, even without action-grounding. If I tell it "Socrates is a man" and "All men are mortal", it correctly infers "Socrates is mortal", not by manipulating meaningless symbols, but presumably by having learned the inferential role of "all", "is a", and the subsumption relation. This is not nothing. While this is not the same as the squirrel's embodied knowledge, it might be a distinct kind of competence worth recognizing as what some might call inferential concepts (Brandom 1994,2008, Bender et al 2021).

Inferential concepts grasp logical and linguistic relationships between terms without grasping the perceptual or causal structure those terms are supposed to track. This makes them powerful for certain tasks (formal reasoning, linguistic manipulation, pattern recognition in text, and translation) while fundamentally limited for others, such as novel physical reasoning, genuine world modeling, or tasks requiring perception-action loops.

Transformer attention computes context-sensitive relationships within a very high-dimensional embedding space whose geometry reflects the inferential regularities present in its training corpus. Attention weights reflect how strongly each token contributes to the context-sensitive structures of the embedding space, which are then used to compute next-token probabilities. Rather than storing explicit propositions or symbolic rules, LLM models acquire a distributed organization that supports systematic generalization, analogy, and logical inference over linguistic expressions. The model's latent space is a compressed geometry of entailment and exclusion relations (what follows from what, what rules out what) that determines how terms like "all," "is a," and their surrounding vocabulary pattern together. These structures are not action-grounded: they are shaped by patterns of language use rather than by successful engagement with the environment. They are, however, richly inferential, encoding how expressions constrain and support one another within historically evolved linguistic practices.

The squirrel's engagement with branches is continuously constrained by success and failure in the physical world. Its action-grounded concepts are therefore directly calibrated through embodied interaction. An LLM's inferential concepts, by contrast, are calibrated through the statistical regularities of linguistic practice.

## 2.5 Metaphoric Extension

According to Lakoff and Johnson (1980, 1999), inferential organization does not arise independently of bodily engagement: abstract domains inherit much of their entailment structure from sensorimotor schemas through systematic metaphorical extension.

Time, for instance, is routinely understood in terms of spatial traversal. We speak of “approaching deadlines”, “leaving the past behind”, “looking forward to the weekend”, or “putting difficulties behind us”. Events are said to “come up”, “pass by”, or “lie ahead”. These expressions are not merely figures of speech: they reflect a deeper cognitive mapping in which temporal reasoning borrows its inferential structure from bodily movement through space.

Quantity is structured in a similar way through vertical spatial schemas. Prices “rise” and “fall”, numbers “go up”, temperatures “drop”, and economic growth “climbs” or “declines”. The mapping associates “more” with “upward movement” and “less” with “downward movement”, mirroring bodily experiences such as stacks becoming taller as objects are added.

Causation, likewise, is frequently conceptualized through the dynamics of physical force. We speak of “pushing someone into a decision”, “driving change”, “triggering a reaction”, “forcing an outcome”, or “being under pressure”. Causal reasoning thus inherits the structure of bodily interactions involving pushing, resistance, momentum, and impact. In each case, the inferential relations among abstract concepts are scaffolded by patterns originally learned through action in the physical world.

On this view, the inferential competence of human cognition is parasitic on action-grounded competence in a precise sense. The host—the embodied agent—can survive and function without deploying inferential, linguistically articulated abstractions. Our squirrel illustrates this vividly: it navigates a complex physical and social environment with considerable sophistication, relying entirely on action-grounded concepts that are directly answerable to the world. An LLM, by contrast, cannot exist without the host. It draws its structure, content, and epistemic grip on reality from the action-grounded layer it extends.

LLMs are trained on the linguistic outputs of grounded agents and therefore inherit the surface patterns of metaphorical extension. They can fluently reproduce expressions such as “rising prices”, “approaching deadlines”, or “forces driving change”. But reproducing the linguistic traces of metaphorical mappings is not the same as participating in the embodied interactions that originally gave those mappings their structure.

## 2.6 Referential Profiles

While Frege and Searle ground reference (and thus meaning) as descriptive clusters; for Kripke proper names are grounded through rigid, extrinsic causal hookups. Rhee (2026), Eliasmith (2013) and others reject the naive "hooking onto the world" and replace it with a rich, distributed, high-dimensional vector structure rather than a symbolic tag. Rhee (2026) suggests grounding reference as public, norm-governed pattern of use, a discourse-level structure that supplies rigidity and stability without causal metaphysics. According to Rhee (2026), for words to refer they must be grounded in what he calls *referential profiles*:

> A referential profile is a structured set of descriptions, memories, perceptions, inferential routes, affective orientations, social uses, and correction practices through which an expression, sentence, or discourse can converge on an entity, situation, relation, or domain. Human beings realize such profiles through perception, episodic memory, embodiment, affect, and social responsibility. LLMs do not. (Rhee 2026)

Referential profiles capture the whole pattern of correct/incorrect continuation built up across a discourse community's history of use. A situational instantiation of this broader profile gets activated for a given expression in a given context via context-sensitive computation over weights and attention. LLMs inherit, according to Rhee, referential profiles from human practice, sedimented in language. In a linguistic format these recurrent relations can be transformed into distributed vector structures.

Although LLMs reproduce the inferential and metaphorical patterns present in human discourse with remarkable fluency, they do not participate in the perception–action loops through which those patterns acquire their normative grip on the world. What appears as participation in linguistic practice is therefore structurally asymmetric: the human interlocutor contributes the grounding conditions that sustain meaning, while the model merely manipulates the linguistic traces of those grounded representations. Linguistic practice can transmit content across agents who already possess it; it cannot bootstrap content in systems that lack the underlying action architecture.

## 2.7 Affordances

Affordances are possibilities for action provided by the environment, i.e., relations between aspects of the material environment and abilities available in a form of life Skilled intentionality is the organism's tendency toward an optimal grip on multiple relevant affordances simultaneously. Rietveld & Kiverstein (2014) distinguish between a landscape of affordances, i.e., “the whole spectrum of abilities available in our human socio-cultural practices” and the field of relevant affordances, which stand out as relevant for a particular individual in a particular situation (Bruineberg & Rietveld 2014).

### Affordances as Control Structures

In contrast to Rhee, Bruineberg & Rietveld's account is explicitly anti-representational: an affordance is picked up by an agent without first encoding it representationally. Rather, what is often described as representation-based cognition is replaced by temporally extended skillful coordination within richly structured landscape of affordances. Perception, accordingly, does not involve representational content but is instead understood as a skilled activity of engaging with multiple affordances. The re-enactment of these percepts during counterfactual thought or reasoning does not become representational either, just by virtue of being re-enactments.According to Kiverstein & Rietveld there is no basis for positing a sharp conceptual divide between online and offline cognition:

> We should not think of offline cognition as a distinct type of cognition, but as a more complex form of coordinating nested states of action readiness and activities to multiple relevant affordances. Such a process is complex because of the nesting of the activities and their increasing reach through time. (Kiverstein & Rietveld 2018, 157)

Rather than representing a sentence internally, a translator coordinates multiple affordance landscapes across different time scales, such as typing, establishing, evaluation and maintaining syntactic expectations, anticipating later discourse, remembering terminology, etc. These nested action-readiness states (see Table 1) can be experienced as bodily affective tension that motivates a skilled individual to act on particular possibilities for action. The layers and states do not evaluate against stored representations, they are assessed as situated appreciation expressed in normative behaviour.

From this perspective, affordances can be understood to function as control structures rather than representations. They do not encode detached descriptions of the world but organize the dynamics of perception, attention, and action by constraining the range of behavior that is available to an organism in a particular context. A branch affords grasping only for an organism capable of climbing; a sentence affords continuation only for a speaker embedded in a linguistic practice. Affordances offer a situated normativity that is world-answerable on a gradual rather than bivalent True/False evaluation schema. Situated normativity is the ability of skilled individuals to distinguish better from worse, adequate from inadequate, appropriate from inappropriate, or correct from incorrect in the context of a particular situation. Affordances therefore function as relational constraints that channel behavior toward some trajectories while suppressing others.

Table 1: Embodied action and increasingly abstract layers of generalization

| Layer | What is being coordinated | Action-readiness to: |
|---|---|---|
| **Sensorimotor coupling** | Perception ↔ movement | Perform an immediate bodily action |
| **Action-grounded concepts** | Affordance patterns generalize across variable situations | Act in a recurrent way to stabilize a class of possibilities |
| **Inferential concepts** | Relations among action possibilities | Coordinate stabilized possibilities with other possibilities |
| **Propositional thought** | Linguistically articulated configurations of affordances | Act on/about a linguistically articulated situation |
| **Reflective metarepresentation** | One's own/another's cognitive activity | Monitor, revise, justify, or regulate cognition |

While humans engage in nested fields of enlanguaged affordances, LLMs model the linguistic traces of human affordance-responsiveness and re-enact linguistic practices. Within its probabilistic latent structure LLMs learn among other things, 1) inferential expectations, 2) discourse conventions, 3) genre regularities, 4) pragmatic patterns. However, as affordances are normally considered relations between an environment and the abilities of an agent, and LLMs lack the relevant embodied abilities to engage with its environment, LLM translation becomes the re-enactment of historically sedimented linguistic behavior that results from enlanguaged affordances

## Enlanguaged Affordances and Reference

Kiverstein & Rietveld (2018, 2020) introduce the notion of *enlanguaged affordances* which likewise can be understood as higher-order control structures: historically stabilized constraints that organize appropriate linguistic continuations, interpretations, and corrections within a discourse community. Kiverstein & Rietveld suggest an anti-representational framework of language use "without making any appeal to internal, content-carrying representational states".

Despite their non-representational nature, enlanguaged affordances carry reference. Reference emerges when historically stabilized practices allow to a pick an enlanguaged affordance that can be flagged same across different speakers, different phrasings, a later moment in a conversation, and can be challenged, corrected, and reaffirmed. This is identical to Rhee's referential profile, a specific, norm-governed pattern of agreement, correction, and re-identification that keeps a name or description anchored to one thing rather than another across occasions.

On a representational picture, a term refers by standing in a checkable relation to an object, held fixed independently of use and consulted when needed. On the control-structure picture, there is no such standing relation to consult. What exists instead is a distributed disposition of the discourse community itself: a constraint that channels which continuations, corrections, and challenges count as apt responses to a given use, and suppresses those that don't. A term "refers" to an object not by pointing at it via some intermediary content, but by being the thing around which this behavior is organized: reference is constituted by the shape of the constraint, not tracked by consulting a separate representation of what the constraint is about. This is why enlanguaged affordances can do referential work while remaining non-representational in Kiverstein and Rietveld's sense: the control structure doesn't encode the fact that "Aristotle" picks out a particular person and then get consulted; it simply is the historically stabilized pattern of correction and agreement that keeps a discourse community's use of "Aristotle" convergent, and reference just is what that convergence, sustained over time, amounts to.

Reference draws on both the action-grounded and the inferential layers, without being reducible to either. From action-grounded, occasion-bound affordance engagement, it inherits its anchor: the initial rigid pick that gives the practice something determinate to track in the first place. From inferential concepts, it inherits the machinery of identity-tracking that lets a discourse community recognize when two paraphrases commit to the same thing. Neither ingredient is sufficient alone: occasion-bound anchoring without inferential identity-tracking gives only isolated, unconnected picks with no way of recognizing sameness across reformulation; inferential organization without an anchoring engagement gives only internally coherent substitution patterns with nothing to substitute about. Reference is what results when a discourse community's inferential competence for identity and substitution is exercised on picks that are, at least at some point, anchored in genuine occasion-bound engagement, and stabilized as a corrigible practice over time.

### Emergence of Representation

Through recurrent interaction, agents in a discourse community learn to coordinate affordance patterns across situations. When these patterns become sufficiently stable, they support a first, pre-propositional form of stabilization: an anticipatory pattern that can be re-presented in perception, memory, or imagery even when the affordance itself is not immediately available. This extends an action-relevant organization of possibilities beyond the immediate situation, but it remains tied to the individual engagement that produced it and does not yet secure representation in the fuller sense. Action-grounded concepts are necessary for representations to evolve because they provide the initially stabilized, action-relevant content that can later be abstracted from immediate engagement and made available for discursive articulation. Without such grounding, there is no independently stabilized content for representation to pick out, manipulate, or redeploy.

A fuller sense of representation requires the community to develop a further specific discursive capacity: propositions (see Table 1), that is, an enlanguaged affordance of asserting-and-being-held-accountable, creates discrete slots, positions that can be affirmed or denied. Only once these slots exist can representations be detached and reused. Representations, in this full sense, are the capacity to treat what fills the propositional slot as a portable object, separable from the specific occasion and speaker who first asserted it. Representations are then available to be negated, embedded, quoted, or redeployed elsewhere. On this view, affordances ground representation at both levels, pre-propositional anticipatory stabilization and full discursive representational detachment. But only the second level constitute representations in the classical sense, since what gets detached and redeployed is not a raw affordance-pattern but an “assertoric commitment” (Brandom 1994), already discretized into something that can be true or false.

Note that this reverses the conventional representational view in which representations are considered basic entities that fill propositional slots. Here, propositionality gets first constituted through enlanguaged affordances. Representational content is therefore not presupposed but emerges when socially stabilized affordance patterns become discretized as assertible and accountable commitments within linguistic practice.

## 2.8 Relevance Realization

Representations and propositional structures are not basic entities but learned forms of stabilized generalization that emerge through interaction with the environment. Vervaeke et al. (2012) introduces the notion of Relevance Realization (RR) as a process that precedes language, concepts, and propositional thought. RR addresses the relevance problem: at any moment an organism is confronted with effectively infinite environmental features, yet it must rapidly determine which of them are relevant for ongoing action. Because exhaustively evaluating all possibilities that the environment offers at any moment is computationally intractable, cognition cannot represent and process every aspect, as a representational account would require.

Although related to Harnad's (1990) symbol grounding problem, the relevance problem is conceptually distinct. Whereas Harnad asks how symbols acquire meaning, RR provides a mechanism to explain how potentially meaningful information becomes salient in the first place. To overcome, among other things, the symbol-grounding problem, Vervaeke proposes that relevance is enacted through embodied, goal-directed engagement with the world. Within this framework, relevant affordances are selected dynamically through successful co-constitution of the agent and its environment. RR explains how an organism selectively attends, stabilizes, and exploits those action possibilities, thereby continuously shaping the affordance landscape by weighing and (re)organizing available opportunities for adaptive behavior. Cognition develops as progressively more sophisticated forms of coordination and organization through increasingly abstract layers of generalization and reasoning. Table 1 shows five nested scales of action-readiness, ranging from immediate sensorimotor responsiveness to increasingly temporally extended and reflexively organized forms of activity.

Vervaeke maintains that many animals participate in relevance realization without possessing representations in the classical sense. Animals clearly engage in sensorimotor coupling, and many mammals (as our squirrel) and birds plausibly possess action-grounded concepts that support flexible and adaptive behavior. Humans, in contrast, exploit the full hierarchy, including inferential, propositional, and metarepresentational forms of engagement.

The taxonomy in Table 1 suggests, on the one hand, that action-grounded concepts need not yet be inferential. For example, a dog may possess an action-grounded concept such as *caregiver*, enabling recognition and appropriate interaction, without representing the rich inferential network that humans can articulate around caregiving, responsibility, ownership, and social norms. On the other hand, the distinction between inferential organization and propositional thought has implications for translation, suggesting that successful translation does not necessarily require propositional thought. Translation may instead depend on inferential organization, i.e., stable networks of linguistic regularities that constrain how expressions relate to one another across contexts.

Meaning, on this account, is grounded in relevance realization within an embodied landscape of affordances and becomes stabilized through recurrent action-grounded engagement within socially coordinated linguistic practices. Kripke's account of initial baptism and the subsequent causal-historical transmission of reference provides a useful parallel: reference is not fixed solely through inferential relations among descriptions but depends on historically maintained relations to what linguistic communities have successfully picked out. LLMs do not participate in this grounding process. They build upon recorded traces of human linguistic practices and can reproduce aspects of their inferential organization without participating in the affordance-generating activities through which those practices acquire meaning.

It is consequently doubtful that LLMs engage in propositional thought in the full agentive sense, even though they can instantiate sophisticated linguistic organization. They can reproduce entailment relations, assess internal coherence and exploit statistically inherited regularities, thereby approximating aspects of truth through the historically sedimented linguistic practices encoded in their training data without propositions functioning for them as action-guiding commitments within an embodied and affectively modulated affordance landscape.

# 3 Possible Objections

Three related objections might be raised against the foregoing argument. First, grounding might emerge through conversational interaction itself, as humans and LLMs jointly construct increasingly sophisticated lines of reasoning. Second, reinforcement learning from human feedback (RLHF) might appear to provide a form of grounding by continually aligning the model's behavior with human judgments. Third, one might argue that grounding could be achieved by embedding an LLM within a robotic body capable of perceiving and acting in the physical world. Although these objections appeal to different mechanisms—dialogue, social feedback, and embodied sensorimotor coupling—they share the intuition that grounding might arise by enriching the model's interactions beyond text alone. I argue that none of these routes, by itself, is sufficient to establish reference and intentionality in the sense outlined here.

## 3.1 Conversational Interaction

The first objection arises from the phenomenology of conversational exchange. In human–LLM interaction, dialogue appears genuine: the human asks questions, the model responds, humans correct or extend, and the exchange gradually refines the content. One might therefore ask whether this back-and-forth itself provides a form of grounding. The human contributes action-grounded concepts, the model responds in ways shaped by them, and the resulting exchange can be richer than either participant’s solitary output. Does the interaction supply what neither system possesses alone?

I argue that it does not. In such exchanges the human performs all the grounding work. The human introduces action-grounded concepts, deploys them in questioning, evaluates responses against them, and uses the model’s output to extend their own thinking. The model functions as a highly sophisticated mirror, reflecting the structure of the human’s concepts in elaborated or reorganized form. Yet it does not acquire grounding from the exchange. Unlike a squirrel, whose action-concepts are continuously updated through perceptual contact, an LLM possesses no persistent, long-term states that are modified by user interaction beyond in-context learning often without modifying its weights[3]. Each conversational turn is processed by the same fixed parameters, encoding statistical patterns in text rather than causal contact with the world. Conversation therefore adds no grounded concepts to the model’s repertoire; indeed, the model has no repertoire in this sense, only a probability distribution over tokens.

This asymmetry becomes clearer when considering what each participant gains. A human may genuinely learn from the exchange—revising concepts, discovering implications, or articulating connections previously unnoticed—because their inferential concepts are anchored in action-grounded ones. For the model, however, nothing is retained: no parameters are updated, no concepts acquired, and the system that processes the next query is identical to the one that processed the previous. The appearance of mutual enrichment is thus misleading: one participant learns, the other merely processes.

Ironically, this helps explain why such systems can seem persuasive interlocutors. Because LLMs are trained on linguistic outputs of grounded thinkers, they reproduce the surface form of grounded inference with remarkable fluency—even arguments about grounding itself. But this fluency reflects the fact that language encodes the structure of embodied experience so thoroughly that its patterns can be reproduced without sharing the experiences that generated them. The map can be copied without consulting the territory.

[3] Not however, that some contemporary systems may also incorporate persistent memory, retrieval, tools, or multimodal input.

### 3.2 RLHF and the Illusion of Grounding

One might object that reinforcement learning from human feedback (RLHF) closes the grounding gap by introducing exactly the kind of external signal that pure statistical learning lacks. In RLHF, human raters evaluate model outputs, and their preferences are used to shape the model's behavior through a reward signal. Does this constitute a form of causal contact with the world, the normative world of human judgment? And if so, does it provide the corrective feedback loop that distinguishes mere symbol manipulation from genuine representation? I will support the first hypothesis and reject the second.

Mollo and Millière (2026) contend that RLHF supplies what they call "world-involving functions", evaluative standards that go beyond intralinguistic prediction, and that these functions are sufficient, in principle, for LLMs to achieve referential grounding in the sense that matters: their internal states acquire correctness conditions relative to how the world actually is, not merely relative to how text tends to co-occur. On their view, the world enters the system not through the eyes or hands but through the normative channel of human evaluation, which is itself anchored in embodied experience.

I'd like to address two points. For one, Mollo and Millière posit "When the model succeeds in generating a factually accurate response, its internal states fulfil their function of representing relevant worldly states." There are several counterexamples. The most striking is probably what Berglund et al. (2023/2024) call *The Reversal Curse*: when an LLM trained on 'A is B' fails to generate 'B is A'. For example, if an LLM can correctly answer "Who is Tom Cruise's mother?" it may fail on the logically equivalent reverse, "Who is Mary Lee Pfeiffer's son?" However, if the model's internal state genuinely fulfilled the function of representing a state of affairs (Tom Cruise's mother is Mary Lee Pfeiffer), that representation should be usable compositionally in either direction. But the fact that accuracy is direction-locked indicates a surface association rather than a state that represents the relation.

The Reversal Curse relates to the discussion on rigid designation above. If "Tom Cruise" rigidly designates a fixed individual, and "is the mother of" is genuinely represented as a relation holding between that individual and another rigidly-designated individual (Mary Lee Pfeiffer), then the relation should be equally retrievable from either end, which it is not. A reason for this may be found in the transformer architecture itself. The feedforward layers function like key-value associative memories, and a key-cued lookup is exactly the kind of mechanism that would produce direction-locked retrieval rather than an address-independent representation.

Ouail et al. (2024) suggest that the *Reversal Curse* can be weakened when construed as a factorization problem: the difficulty may arise because the model learns a relation under one factorization of its arguments but fails to recover the same information under another. However, eliminating this factorization asymmetry would show only that the information is accessible under alternative retrieval paths. It would not show that the model represents the underlying relation R or the inverse relation $R^{-1}$ with its ordered arguments as such. Even a complete solution to the factorization problem would therefore not by itself establish that the model possesses relational representations. Similar objections have already been put forward by Fodor and Pylyshyn (1988) in their criticism of connectionism.

A second point is that Mollo and Millière seem to mistake RLHF training as the transmission of a grounding relationship for the possession of one. The objection is structural: RLHF introduces a signal that remains entirely within the linguistic domain. Human raters evaluate text against text; they judge whether a generated response is coherent, helpful, etc. The feedback loop runs from language, through human judgment, back to language. The world beyond language is never directly implicated in what is transmitted to the model. A rater expresses a preference grounded in their own embodied experiences and concepts, acquired through years of embodied engagement with the world. But what is transmitted to the model is not that grounding; it is only a scalar encoding surface preferences over strings. The model learns, in effect, to produce outputs that human readers find plausible. It learns the shadow of grounded concepts without acquiring the substance.

This distinction maps on the notions of direct and derived proper function already present in Millikan (1984). A direct proper function is fixed by a device's actual track record: a representation succeeds when it reliably tracks the environmental condition it was selected to track, independent of anyone's judgment. A derived proper function, by contrast, is assigned by the intentions of those who designed or use the device. Success here means satisfying what the designers or evaluators intended, whether or not the world cooperates (see Wheeler's *design-intention*).

RLHF trains toward the second standard: it rewards outputs that human raters judge appropriate, not outputs that are independently verified against the world. Factuality is measured against what a human rater, working from memory and limited time, judges to be factual: "Helpfulness" is glossed as "genuinely assist[ing] users with their goals," but the training signal is a rater's prediction of helpfulness, not a verified downstream outcome. "Harmlessness" is glossed as requiring "sensitivity to causal relationships between linguistic outputs and potential worldly outcomes," but what is rewarded is a rater's imagined causal story, not an observed one.

Mollo & Millière describe here a training process that shapes the model's internal organization to fit a structured field of human evaluative practice. This is not a causal-informational relation to worldly states of affairs. It is training towards a normatively structured, socially instituted field of correction, endorsement, and appropriate continuation, responsive to a well-targeted field of enlanguaged affordance. However, it is lacking sensorimotor and action-ground components. RLHF, on this reading, does not endow the model's internal states with causal-informational content and “intrinsic meaning”; it sculpts the model's latent organization to track the contours of an affordance landscape constituted by evaluator’s norms. Mollo & Millière’s argument, intended to secure world-answerable grounding, in fact provides independent support for the enactive alternative developed here.

The reframing also predicts training signals in closed, non-human-mediated loops, such as for instance verifiable mathematics, executable code with passing or failing tests, tool calls returning real errors. Here, the reward is fixed by an actual outcome, not by a rater's guess, which brings it closer to Millikan's direct proper function. The enactive account anticipates that world-answerable (propositional) content becomes available precisely where genuine, closed-loop coupling with the world is present. It remains unavailable where training closes its loop with human judgment instead. Mollo & Millière's strongest cases and this paper's predicted exceptions are, tellingly, the same cases. However, even RLHF does not endow LLMs with action-grounded concepts.

### 3.3 Grounding LLMs in a Robot

An LLM embedded in a robotic body equipped with cameras and actuators would overcome the most basic limitation identified by Harnad’s symbol grounding problem: it would no longer be confined to linguistic relations but could enter into direct sensorimotor coupling with its environment. It could visually detect branches, attempt to grasp them, succeed or fail in navigating obstacles, and adapt its behavior accordingly. Such a system would participate in causal chains linking internal states to worldly events. However, causal embedding alone would not yet establish reference in the robust sense required here. Genuine functional and normative organization requires the loop to close with the world itself, not with another model's judgment about it.

For reference to emerge, environmental coupling must become integrated into a functional and normative organization of behavior. On Dretske’s (1981) account, a state represents something when it carries information about it through reliable covariance that is recruited for behavioral control. On Millikan’s (1984) account, representation requires a stabilizing function: internal states must acquire the role of tracking particular environmental features, such that error and misrepresentation are defined relative to that function. The ecological-enactive perspective developed here adds that intentionality consists in skilled responsiveness to relevant affordances, while reference emerges from the public stabilization of such intentional organization within shared practices.

Accordingly, grounding is not achieved merely by adding perception and action to a statistical language model. What is required is the integration of sensorimotor coupling into a system capable of distinguishing successful from unsuccessful engagement, maintaining functional norms, and participating in practices where uses can be corrected and stabilized over time. This stabilization must also remain ongoing rather than merely historical. A system trained offline and then deployed with frozen weights inherits a function fixed at training time but loses any mechanism for detecting when that function has drifted from present circumstances. Only a system whose selection process remains open at deployment — through continual or online learning from real-world consequences — secures function that is answerable to the world on an ongoing basis, rather than merely once, historically, in the past.

A robotically embodied LLM might therefore acquire causal grounding without yet acquiring reference in the full human sense. Genuine reference requires not only a world-model but a way of being answerable to the world through embodied relevance realization, functional normativity, and participation in the social practices that make correctness and error intelligible.

## 4 How do Humans and LLMs translate

In *Word and Object*, Quine (1960) develops a thought experiment in which a field linguist encounters a completely unknown language. In this famous example, a rabbit hops by and a native speaker says "Gavagai." No amount of behavioral observation, Quine says, even in principle, can definitively settle whether that might mean "rabbit," "undetached rabbit part," "rabbit stage," "lo, rabbithood is instantiated," or something entirely else. The same observable behavior is compatible with radically different ontological schemes and referential semantics. Quine concludes that there are infinitely many incompatible “translation manuals” (i.e., systematically organized sets of rules for mapping expressions from one language to another) that are equally consistent with all possible behavioral evidence. Thus, Quine concludes, translation is radically underdetermined by all possible evidence.

If this is so, LLMs seem to be in an even worse epistemic position than Quine's field linguist:

1. The field linguist has at least perceptual access to the world the speakers are talking about, they can see the rabbit, observe what natives do in response to rabbits, etc.
2. The LLM has only text, no perceptual grounding, no behavioral evidence, no ostension, no causal contact with referents.

Yet LLMs translate between languages with surprising accuracy. How is this possible if, as Quine says, no finite behavioral evidence uniquely fixes reference, if translation is radically indeterminate?

## 4.1 Geometry of Embedding Spaces

One of the most striking empirical findings is that the embedding spaces of different languages become approximately isomorphic when models are trained on multilingual data. These embedding spaces can be exploited for translation under some conditions but fail under others, especially for typologically distant and low-resource languages. This was first discovered in Word2Vec (Mikolov 2013) and has been confirmed repeatedly in transformer models. Carl (2021), for instance, shows that is possible to find a linear transformation (a rotation matrix) that maps English embeddings (Glove) onto Spanish embeddings (SBW-vectors-3008) with good accuracy for high frequency words, even though no such alignment was ever explicitly trained, other than with a small bilingual seed lexicon.

Multilingual LLMs are usually also not trained on explicitly / purposely aligned bi- or multilingual data, but they do not use a bilingual seed lexicon. However, the internet contains enormous amounts of *incidental* parallelism — Wikipedia articles on the same topic in different languages, multilingual websites, news agencies that publish in multiple languages, academic papers with multilingual abstracts. This is not parallel data in the explicit sentence-aligned sense, but it provides document-level co-occurrence of the same content across languages, which is a weaker but still meaningful alignment signal.

Some researchers argue this incidental parallelism does most of the heavy lifting that provides LLMs the ability to translate (Wang et al 2024; Qorib et al 2025). Others argue the language geometry argument is primary (Artetxe et al. 2018). In their view, multilinguality arises because languages are mapped into a shared latent manifold shaped by distributional similarity, but the incidental bilingual snippets are not required to explain the core phenomenon. Balashov (2025) claims that translation abilities in LLMs arise from both, continuous interaction between local bilingual clues and global statistical alignments.

Several properties of transformer training amplify this implicit alignment:

- **Subword token sharing.** Romanized loanwords, proper nouns, numbers, and punctuation are often encoded by identical or overlapping character sequences across languages. "Paris", "Tokyo", "DNA", "COVID", "2024" appear across dozens of languages with identical or near-identical token sequences. These shared anchors act as *Rosetta Stone fragments*, pieces of bilingual corpora or bits of parallel data that serve as ‘keys’ to cross-lingual alignment, which are scattered throughout the training data. Such Rosetta Stone fragments force the model to learn mappings of collocations across languages because the same token must serve coherent roles in all of them.
- **Shared positional and syntactic structure.** Many languages share broad syntactic patterns — SVO ordering, subordinate clause structures, question formation — and the attention mechanism learns these patterns in a way that generalizes. The model discovers that certain structural relationships recur across languages, which build cross-lingual grammatical intuitions without explicit pairing.
- **Transformer Architecture.** Research has shown that activation patterns in the deeper layers of the transformer architecture become more heavily contextualized, so that they increasingly encode language agnostic structures rather than surface form. Translation can then be explained as a decoding from this ‘universal’ space of activation structures into the target language surface form.

## 4.2 Where implicit Training breaks down

The implicit alignment isn't perfect, and the failures are instructive:

- **Low-resource languages** that appear rarely in training data have poorly shaped embedding spaces — the model sees too few examples to learn the full relational geometry. Translation quality degrades sharply for languages like Yoruba, Swahili, or Lao compared to French or German, roughly proportional to how much data existed in training.
- **Language-specific concepts** that have no real equivalent in other languages may be handled poorly. Japanese "木漏れ日" (the interplay of light and leaves), German "Schadenfreude", or grammatical features like grammatical gender, honorifics, or evidentiality markers (which encode *how you know* something is true) have no anchor in languages that lack them. The model may drop or approximate them with a phrase, paraphrase, or construction in translation.

- **Register and pragmatics** are often lost. Formal versus informal address (French tu/vous, Japanese keigo) requires cultural knowledge about the social relationship between speakers that a statistical model only partially captures.

How does this answer Quine's indeterminism problem? How do LLMs logically solve the indeterminacy of translation?

### 4.3 Inferential Concepts as the Basis of Translation

Quine's (1960) thesis of the indeterminacy of translation establishes an important, logical point: No amount of linguistic evidence alone uniquely determines a single semantic interpretation or underlying ontology:

Distributional similarity ⇏ Semantic similarity

Quine takes this to imply that there is no uniquely correct translation grounded in determinate reference. As discussed in section 1, Quine's indeterminacy thesis also stands in direct opposition with the distributional fallacy, The success of contemporary LLMs, trained on huge corpora and used for machine translation, suggests that either Quite was wrong and semantic meaning can indeed be reconstructed from linguistic distributions, or that translation need not recover a uniquely fixed semantic or referential structure. I argue for the latter: Quine's underdetermination need not undermine translation itself, rather translation may succeed by preserving sufficiently similar inferential, pragmatic, and behavioral organization across languages. This suggest a third version of the DH, according to which distributional regularities can be exploited to predict subsequent linguistic behavior without first being converted into determinate semantic representations:

Distributional regularities ⇒ Predictive linguistic organization

Thus, Quine's logical point remains intact, but its implications for translation appear considerably weaker than he supposed. Davidson (1984, 2001) reverses Quine's skeptical indeterminism, suggesting that there are always different ways of representing the same meaning. Davidson introduces a *principle of charity* suggesting that an interpreter constructs a theory of meaning that narrows the space enough that practical interpretation becomes possible. The charity principle presupposes that we must assume most speakers' beliefs are true, that their reasoning is broadly coherent, and that their utterances fit into a shared world – thus interpretation is not radically unconstrained. Malmkjær (1993) takes this to defend the possibility of translation.

In addition, as Tymoczko (1999) argues, translation is constrained by historically, culturally, and intertextually available knowledge, such that not every interpretation constitutes a viable translation. At the same time, she recognizes that translators must make choices among underdetermined possibilities, thereby locating indeterminacy not simply as a failure of translation but as a source of translator agency. Thus, one possible TS response is that Quine's indeterminacy applies to an artificially impoverished epistemic situation, not necessarily to situated translation practice.

LLMs, however, do not participate in a situated shared world but they are very good in modelling the linguistic behavior of situated concept-users, without themselves being concept-users in the full sense. LLMs are extraordinarily sophisticated radical interpreters. LLMs can be seen to model the inferential structure that charitable interpreters attribute to speakers and thereby approximate concept use. They capture the patterns of entailment, association, and substitutability that concepts participate in, without capturing the referential content of concepts. This is sometimes called capturing the inferential role semantics while missing the truth-conditional semantics (Block 1986).

When a model is trained on a large multilingual corpus, it sees text in many languages describing the same world. The word "dog", "Hund", "chien", "كلب" and "犬" all appear in contexts involving the same real-world concepts, they co-occur with words for barking, leashes, parks, veterinarians, etc.. Even without ever seeing a sentence paired with its translation, the model learns that these words occupy similar positions in their respective collocational spaces: if words from different languages share structurally comparable contextual distributions that occupy geometrically similar positions in a shared collocational space they can be treated as translations of one another. A similar assumption has been labeled "formal translation correspondence" in a structuralist context and defined as "any TL category (unit, class, structure, element of structure, etc.) which can be said to occupy, as nearly as possible, the 'same' place in the 'economy' of the TL as the given SL category occupies in the SL" (Catford 1965: 27).

On this view, Translation does not require that the translator—or an LLM—recover the referential content of a source text. Rather, successful translation depends on preserving the inferential relations that concepts bear to one another across languages. LLMs achieve this by learning a multilingual space of inferential correspondences, in which lexical items occupy analogous positions despite lacking direct action-based grounding in the world. Translation thus becomes the realization of structurally equivalent inferential organization rather than the transfer of fully articulated semantic representations.

### 4.4 The representational Bias in Translation Studies

Clark and Toribio (1994) characterize certain cognitive activities as representation-hungry. Representation-hungry tasks, they say, cannot be solved through direct sensorimotor coupling with the current environment alone, but require reasoning about absent, hypothetical, counterfactual, or abstract states of affairs. These forms of cognition appear to require internal representations, with determinate content that can be manipulated independently of ongoing perception and action.

Translation has traditionally been regarded as a representation-hungry task (e.g., Interpretive Theory, Seleskovitch & Lederer 1984; Relevance Theory, Sperber Wilson 1986/1995). Classical theories of Translation Studies (TS) assume that translators and interpreters first construct an internal representation of the source text's meaning before reformulating that representation into the target language (Gile 1995). The representational commitment in TS also runs, more quietly, through the functionalist and norm-oriented paradigms that have long presented themselves as alternatives to equivalence-based models. Adequacy is obtained or assessed via an intermediate, portable, decontextualized representation (kernel meaning, functional profile, norm, skopos, strategic stance) that mediates between ST and TT and can be entertained, stored, and manipulated independently of the immediate reading of the source text.

Nida's (1964) dynamic equivalence presupposes a decomposition of the source message into abstract, culturally-neutral "kernel" propositions — a deep-structure-like representation, indebted to the generative grammar of the period — which is then restructured for the target reader; equivalence is achieved at the level of this intermediate representation, not at the level of surface form.

Theories of equivalence assume that, although languages differ and perfect equivalence may be impossible, translators can identify sufficiently stable semantic/pragmatic correspondences. Translators do not require logical uniqueness, as Quine suggested, they require a sufficiently appropriate translation for a particular communicative situation. Translation can succeed without there being one metaphysically privileged semantic representation.

Vermeer's Skopos theory (Reiß & Vermeer 1984) makes the representational commitment explicit in its very terminology: the skopos is a represented goal-state against which translational action is planned and evaluated, and the translatum is produced by matching output to this prior goal-representation. Nord's (1997) translation brief formalizes this further, requiring translators to construct an explicit representation of intended function before production begins. Also House's (1977/1997) model of translation quality assessment operationalizes a related move: source and target texts are analyzed in terms of functional-pragmatic profiles (field, tenor, and mode), and translation quality is assessed by comparing these profiles rather than by treating the texts themselves as the primary objects of comparison. This shifts translation away from identity of meaning toward functional adequacy. Translation success dependent on situated communicative purposes, not on recovering a uniquely determined semantic object.

Even Venuti's (1995) domesticating/foreignizing distinction, framed as an ideological and strategic rather than a cognitive claim, presupposes that translators represent a global stance toward the source culture and apply it consistently across choices, rather than letting choices emerge locally from engagement with each textual moment.

In Toury's (1995) approach to Descriptive TS (DTS), descriptive norms play a structurally similar role: norms are internalized regularities that constrain what counts as an acceptable solution within a given socio-cultural configuration. From a cognitivist perspective, such norms can be construed as standing representations of translational expectations, tacitly informing the selection among alternative renderings. Similarly, Hermans's (1999) descriptive and systemic approaches constitute a significant sociocultural departure from classical individualistic cognitivism but retain a cognitivist explanatory residue insofar as translational behavior is explained through norms, expectations, decisions, and interpretive structures. DTS thus successfully externalizes many of the determinants of translation while leaving open the question of how these socially stabilized determinants become effective in situated translational activity—and whether this mediation requires internal representation at all.

DTS shifts the object of explanation from semantic identity to socially stabilized translational practice. Translation does not require uniquely determined meaning; translators operate within historically and culturally stabilized constraints on what counts as an acceptable solution. But it is unclear how those norms become effective in the translator: If they are represented internally as expectations or rules, the representational problem has merely moved from semantic content to norms.

Gutt's (2000, 2005) relevance-theoretic approach goes a step further. Here, translation is not necessarily the reproduction of the same semantic content but the production of an utterance that provides an interpretive resemblance to the source. This constitutes a partial departure from representationalism while retaining an inferential/cognitivist architecture. Carl (2025b) argues that Gutt's (2005) distinction between stimulus mode (S-mode) and interpretive mode (I-mode), marks an important departure from classical representationalism. By shifting the focus from the transmission and recovery of encoded meanings toward the preservation of inferential relations and the communicative effects of an utterance, Gutt's account opens a path toward a more process-oriented conception of translation. Nevertheless, Gutt's departure remains deeply anchored in classical cognitivism. What remains underdeveloped, therefore, is the possibility that interpretive resemblance could emerge not from the preservation of representational content, but from the coordination of embodied, affective, and action-oriented processes within a shared communicative environment.

A notable move beyond this residual cognitivism can be found in Robinson (2020, 2023), who explicitly reframes translational norm theory through 4EA (Embodied, Embedded, Enactive, Extended, and Affective) cognition. Rather than treating norms as pre-existing representations that translators internalize and subsequently apply, Robinson locates their formation in embodied, affective, and socially situated interaction: repeated encounters become stabilized as patterns that orient future action, often without ever being explicitly verbalized. In this sense, norms need not be representations that mediate between social structure and translational behavior; they can be understood as emergent, affectively charged orientations to action. Robinson thus points toward precisely the enactive reconceptualization of translational norms that DTS leaves implicit.

### 4.5 Representationalism in Cognitive Translation Studies

Similar representational assumptions underly much of Cognitive Translation Studies, where the vocabulary of "mental representation" is largely taken for granted rather than argued for. Halverson's (2010, 2017) work on cognitive linguistics and translation shift explains regularities in translational choice by appealing to differential entrenchment of construal options as mental representations across the two language systems. Alves's psycholinguistically-oriented models of the translation process (e.g., Alves & Gonçalves 2013) similarly frame expertise in terms of the efficiency with which source representations are constructed and monitored during processing, drawing on relevance-theoretic notions of processing effort. Risku's earlier work (2005) likewise modeled translation as complex, representation-based problem-solving, although her subsequent situated and ethnographic studies (e.g., Risku 2021, 2026) push toward a more distributed, artefact- and network-mediated picture of translatorial cognition. Translation is not performed by a detached observer trying to infer a uniquely determined meaning from linguistic evidence. It is performed by an embodied agent already embedded in a social, material, linguistic, and cultural environment.

Schaeffer & Carl (2013) propose a representational monitor model of the translating mind which is, however, successively re-interpreted and re-mapped in enactivist terms (Carl 2023). The ABC framework (Carl 2025b) takes this enactivist turn further by treating translation not as the manipulation and monitoring of internal representations, but as the dynamic coordination of affective, behavioral, and cognitive processes in situated activity. On this view, cognitive organization emerges from patterns of action, relevance realization, and interaction with the environment, thereby replacing representational mediation with an account of translation as embodied and dynamically enacted cognition.

### 4.6 The Ecological-enactive Perspective

An ecological-enactive perspective offers a way of redescribing these same regularities without positing internal representations that must be constructed and consulted prior to action. On this view, the norms, functional profiles, and skopoi that classical and cognitive translation studies alike treat as represented contents can instead be understood as *enlanguaged affordances* (Rietveld & Kiverstein 2014, Kiverstein & Rietveld 2018): possibilities for translational action that are directly solicited by a socio-material and sociolinguistic landscape — genre conventions, institutional expectations, prior translations, terminology databases, client briefs — rather than internally modeled prior to engagement with it. A skilled translator's sensitivity to what a text "calls for" in a given target context would, on this account, be a form of situated skillful coping with a field of relevant affordances, structured by the practices of a language community, rather than the output of matching a represented source-meaning against a represented target-function. On this account, translators re-enact the relevance-realizing practices through which enlanguaged affordances become meaningful within a target linguistic community. Translation in this ecological-enactive view, consists not in transferring representations but in reproducing appropriate patterns of skilled linguistic engagement under new cultural and linguistic constraints.

## 4.7 LLM Translation beyond Representationalism

From an ecological-enactive perspective, the success of LLMs suggests that the structure required for translation is partially already encoded within language itself, prior to any act of translation. Human languages are not arbitrary collections of expressions but historically evolved systems whose inferential organization reflects centuries of embodied interaction with the world. Large multilingual corpora preserve these regularities across languages and LLMs exploit the stable inferential relations sedimented within linguistic practice.

LLMs may not operate on meaning itself but on the linguistic consequences of meaning. Their internal structures need not correspond to worldly referents, action-grounded concepts, or explicit propositional structures. Instead, they capture the inferential organization that linguistic communities have collectively stabilized through successful communication and encoded in networks of enlanguaged affordances. Translation then becomes possible because different languages have evolved partially corresponding affordance landscapes whose inferential structures reflect broadly shared forms of embodied and cultural engagement with the world. Cross-lingual translation therefore consists not in recovering a unique semantic representation but in identifying functionally corresponding positions within these historically evolved affordance structures.

This is not to say that LLMs are representation-free in the trivial computational sense — their processing is saturated with high-dimensional vectors and internal states that are naturally called representations. The point is rather that these are not representations of the kind the classical picture requires: there is no amodal, propositionally structured, referentially anchored contents that are constructed prior to, and consulted independently of, the act of translating. Rather LLMs seem to operate on compressed, purely relational map of how expressions in one language pattern with other expressions across contexts. On this diagnosis, translation remains a representation-hungry task, as Clark and Toribio's criteria suggest, that can be addressed, however, with statistically sedimented means rather than through amodal symbol-manipulation.

Seen in this light, Quine may have been correct that semantic content remains underdetermined by linguistic evidence. Quine's approach was to ground translation in publicly shared stimulus-meaning, to assent or dissent, rather than in a determinate mentalese. This anticipates in some ways the ecological picture defended here. Successful translation does not require establishing a uniquely correct ontology or fully referential content. It requires producing target-language utterances that occupy sufficiently similar inferential and pragmatic roles within their respective linguistic communities. If alternative translation manuals generate indistinguishable communicative behavior, they are not merely observationally equivalent, but they are translationally equivalent. Table 2 summarizes how different translation theories, explicitly or implicitly, address Quine's indeterminacy.

Table 2: Approaches in Translation Studies to address Quines Indeterminacy hypothesis

| **Approach** | **Response to Quine** |
|---|---|
| **Equivalence** | Meaning is sufficiently recoverable |
| **Functionalism** | Exact meaning is unnecessary; function matters |
| **DTS** | Translation is constrained by norms and conventions |
| **Relevance Theory** | Translation preserves interpretive/inferential resemblance |
| **Situated/4EA approaches** | Translation is constrained by embodied and social context |
| **The ecological-enactive account** | Translation need not reconstruct determinate meaning at all |

LLMs constitute neither a refutation of Quine nor evidence that semantic understanding is unnecessary for human cognition. Rather, they demonstrate that centuries of embodied human interaction have progressively structured language into rich ecologies of inferential relations and landscapes of enlanguaged affordances. LLMs inherit this cultural sedimentation without reproducing the embodied relevance-realizing processes from which it originally emerged. Translation therefore appears to depend less on reconstructing meaning than on preserving the inferential organization through which meaning becomes publicly available in language.

## 5 Conclusion

In his seminal paper Distributional Structure, Harris (1954) states that “If two expressions are semantically similar, they tend to occur in similar collocational distributions.” This captures a semantic-to-distributional relation: semantic similarity is reflected in regularities of linguistic usage. Firth (1957) shifts the direction of inquiry with his dictum that “you shall know a word by the company it keeps,” turning distributional regularities into an epistemological and ultimately engineering principle for identifying and predicting semantic similarity. Distributional methods were subsequently deployed with great success in language modeling, machine translation, and, most recently, LLMs, but were also progressively strengthened into what I call the *Distributional Fallacy*: the ontological claim that distributional representations constitute “full-blown accounts of linguistic meaning” (Sahlgren 2005).

The *Reversal Curse* illustrates a limitation of this stronger claim. Syntagmatic and paradigmatic distributions can encode the linguistic signatures of relations without necessarily encoding the relational structure itself, which is albeit part of linguistic meaning. For example, distributional patterns may capture the directional signatures of $R$=*mother-of(x,y)* and $R^{-1}$=*child-of(y,x)*, yet this does not by itself establish that the model represents $R^{-1}$ as the inverse of $R$, $R(x,y) \Rightarrow R^{-1}(y,x)$.

Thus, LLMs can exploit distributionally organized relational and inferential regularities without those regularities being encoded as determinate relational representations. This motivates a weaker formulation:

Distributional regularity ⇒ Predictive linguistic organization

Predictive linguistic organization, rather than a “full-blown account of linguistic meaning,” may be sufficient for translation. LLMs can exploit the stabilized inferential organization preserved in linguistic corpora, capturing the publicly sedimented consequences of meaning-making without possessing the embodied, action-grounded basis from which human reference and meaning emerge. That is, for humans, relational reversibility, like reference and propositions, is most likely learned from action-grounded engagement (reciprocal role-taking, physically-reversible interaction) rather than a primitive that the mind starts with, as the Language of Thought hypothesis would assume. LLMs trained only on text have no access to that action-grounded channel. The success of data-driven translation is therefore not evidence that linguistic distribution constitutes meaning, but evidence that linguistic corpora preserve enough of the publicly stabilized consequences of meaning for translation to succeed.

This weaker DH also reconciles Quinean indeterminacy with the success of distributional models by separating semantic meaning from linguistic constraint: distributional regularities need not fix reference or constitute semantic content in order to constrain translations or support highly accurate predictions of linguistic behavior. They capture the sedimented consequences of meaning-making without constituting its ground.

Traditionally, translation has been seen as transferring or reconstructing internally represented meaning. It has therefore been considered representation-hungry (Clark & Toribio 1994): successful translation requires determinate representations with genuine referential content. As we’ve seen, LLMs do not provide this possibility, yet they translate and communicate with high accuracy. This paper develops a non-representational, ecological-enactive account suited to explain this puzzle. Rather than beginning with internally represented meanings, it starts from embodied interaction with the world. Through relevance realization, organisms become selectively responsive to enlanguaged affordances, giving rise to increasingly stable action-grounded concepts (including reversible relations) which may eventually stabilize through usage and inference as socially accepted references. Human linguistic communities progressively sediment embodied regularities into systems of inferential relations that preserve the public consequences of successful interaction with the world. Translation, then, operates primarily within this historically accumulated inferential organization rather than by reconstructing presumed underlying mental representation.

Table 3 compares how human and LLMs translation. LLMs translate despite lacking action-grounded concepts, perceptual experience, or referentially anchored propositional content. Their competence derives not from recovering meaning/reference itself but from exploiting the inferential organization that embodied agents have progressively encoded in language. LLMs inherit the products of centuries of human relevance realization without reproducing the relevance-realizing processes that generated them.

The resulting picture in Table 3 places human translation at different levels of the cognitive ecology. Human translators may traverse the full hierarchy from embodied interaction through action-grounded and inferential concepts to linguistic production, while LLMs enter only after this organization has already become publicly available in language. Their success therefore reflects not the absence of meaning from translation but the remarkable extent to which meaning has become historically sedimented in linguistic practice.

Table 3 treats linguistic and inferential structure primarily as sedimentation downstream from embodied interaction. Human concept formation may, however, also involve a reverse direction in which people first acquire words, definitions, and normative uses and subsequently reorganize their experience and action through them. In

Table 3: Human vs. LLM translation. Humans inherit a landscape of enlanguaged affordances through enculturation, while LLMs inherit the statistical traces of historically sedimented linguistic practice. Left: the upward and right arrows indicate cognitive and affective regulation of ongoing translation processes. Relevance realization determines what matters now, Action-grounded concepts stabilize what things are about (reference grounding) and Inferential concepts stabilize how things relate to one another

**HUMAN TRANSLATION**

**Landscape of enlanguaged affordances**
↓
**Field of enlanguaged affordances**

**Source text**
↓
Embodied interaction
↓
Relevance realization
(organization/selection of relevant affordances)
↓
Action-grounded concepts (affordance organization)
↓
Inferential concepts (inferential organization)
↓
**Target text**
↑
Monitoring / Revision / Deliberation
↑
Propositional thought
↑
Reflective consciousness

Affective control, orientation and evaluation → → → →

→

**LLM TRANSLATION**

**Statistical traces of historically sedimented linguistic practice**

**Source text**
↓
:
:
:
:
:
↓
Inferential organization
( Predictive linguistic organization )
↓
**Target text**

the case of rights, obligations, contracts, and social statuses, linguistic practices may constitute the relevant institutional reality.

This perspective resolves two central puzzles discussed throughout the paper. First, it reconciles Quine's indeterminacy of translation with the practical success of both human and machine translation. Quine was right that linguistic evidence alone cannot uniquely determine reference or semantic ontology. However, translation need not recover a uniquely correct referential interpretation. It succeeds by preserving sufficiently similar inferential and pragmatic organization across languages, thereby reproducing comparable patterns of understanding and action within different linguistic communities. Second, the account resolves what was termed the distributional fallacy. Distributional semantics does not constitute a theory of meaning because the statistical regularities in language corpora are the historical products of embodied relevance realization rather than its source. Meaning is not generated by linguistic co-occurrence; rather, embodied interaction with the world progressively structures language into stable inferential relations and networks of enlanguaged affordances.

Because these structures preserve the public consequences of grounded cognition, they provide sufficient information for many translation and other language tasks without requiring the translator—human or artificial—to reconstruct the embodied processes from which they originally emerged.

In this view, enlanguaged affordances are the ecological successor to Searle's descriptive clusters. Searle was right that words derive much of their significance from a rich network surrounding them, but he was wrong, on an enactivist reading, about what that network is: It is not a network of internal descriptions, but a network of publicly stabilized possibilities for thought, discourse, perception, and action. Rather than reconstructing what speakers internally represent, the explanatory focus moves to exploiting what linguistic practices encode.

**Conflict of Interests**: The author declares no competing interests.
**Declaration of funding**: No funding was received
**Use of GenAI**: I have used ChatGPT / GPT-5.6 Luna and Claude Sonnet 5 in this paper for idea generation, idea exploration, language improvement, and copyediting. I have carefully checked the validity and integrity of all GenAI generated content and revised/edited it.

## References

Alves, Fabio; Gonçalves, José Luiz V. R. (2013) Investigating the conceptual-procedural distinction in the translation process: A relevance-theoretic analysis of micro and macro translation units. Target 25/1: 107-124, 2013

Artetxe, M.; Labaka, G.; Agirre, E. (2018) Unsupervised Statistical Machine Translation. In Proceedings of the 2018 Conference on Empirical Methods in Natural Language Processing, Brussels, Belgium, 31 October–4 November 2018; pp. 3632–3642

Baggio, Giosuè and Elliot Murphy. (2026). On the referential capacity of language models: An internalist rejoinder to Mandelkern &Linzen. Computational Linguistics, 1–10.

Balashov, Y. (2025). Translation in the Wild. Information, 16(12), 1077. https://doi.org/10.3390/info16121077

Bender, Emily M. and Gebru, Timnit and McMillan-Major, Angelina and Shmitchell, Shmargaret (2021) On the Dangers of Stochastic Parrots: Can Language Models Be Too Big? Association for Computing Machinery}, New York, https://doi.org/10.1145/3442188.3445922

Berglund, Lukas, Meg Tong, Max Kaufmann, Mikita Balesni, Asa Cooper Stickland, Tomasz Korbak, Owain Evans. (2024) The Reversal Curse: LLMs trained on "A is B" fail to learn "B is A"}, https://arxiv.org/abs/2309.12288},

Block, Ned (1986). Advertisement for a Semantics for Psychology (in Midwest Studies in Philosophy).

Brandom, Robert (1994). Making It Explicit: Reasoning, Representing, and Discursive Commitment. Cambridge: Harvard University Press.

Brandom, Robert (2008). Between Saying and Doing: Towards an Analytic Pragmatism. Oxford University Press

Bruineberg J. and Rietveld E. (2014) Self-organization, free energy minimization, and optimal grip on a field of affordances. Front Hum Neurosci. 2014 Aug 12;8:599. doi: 10.3389/fnhum.2014.00599. PMID: 25161615; PMCID: PMC4130179.

Carl, Michael (2021) Translation Norms, Translation Behavior, and Continuous Vector Space Models. In Machine Translation: Technologies and Applications. Springer, pp 357-388

Carl, Michael (2025a) Representation and Resemblance in Translation: Scrutinizing Interpretive Language Use in Relevance Theory, Routledge (preprint on https://philsci-archive.pitt.edu/24030/ )

Carl, Michael. (2023). Models of the Translation Process and the Free Energy Principle. Entropy 25, no. 6: 928. https://doi.org/10.3390/e25060928

Carl, Michael. (2025b) Temporal Dynamics of Emotion and Cognition in Human Translation: Integrating the Task Segment Framework and the HOF Taxonomy. Digital Studies in Language and Literature, 2025. https://doi.org/10.1515/dsll-2025-0002

Catford, J.C. (1965) A Linguistic Theory of Translation, Oxford: OUP

Chalmers, David (1995). "Facing up to the problem of consciousness". Journal of Consciousness Studies. 2 (3): 200–21

Clark, Andy & Toribio, Josefa (1994). Doing without representing? Synthese 101 (3):401-31.

Damasio, A. (1994). Descartes' error: Emotion, reason, and the human brain. Putnam.

Davidson, Donald. (1984) Inquiries into Truth and Interpretation. Oxford: Clarendon Press..

Davidson, Donald. (2001) Subjective, Intersubjective, Objective. Oxford: Clarendon Press.

De Jaegher H & Di Paolo E (2007) Participatory Sense-Making: An enactive approach to social cognition. Phenomenology and the Cognitive Sciences, 6(4), 485-507.

De Jaegher H & Froese T (2009) On the role of social interaction in individual agency. Adaptive Behavior, 17(5), 444-460.

Dennett, D.C. (1983) Intentional systems in cognitive ethology Behav. Brain Sci.

Dretske, Fred (1981) Knowledge and the Flow of Information. MA: MIT Press.

Eliasmith, C. (2013). How to build a brain: A neural architecture for biological cognition. Oxford University Press.

Evert, Stefan. 2008. "Corpora and Collocations." In Corpus Linguistics: An International Handbook, edited by Anke Lüdeling and Merja Kytö, 1212–48. Berlin: Mouton de Gruyter.

Firth, J. (1957). A Synopsis of Linguistic Theory, 1930-55. In Studies in Linguistic Analysis (pp. 1-31). Special Volume of the Philological Society. Oxford: Blackwell. [Reprinted as Firth (1968)]

Fodor, J. A., & Pylyshyn, Z. W. (1988). Connectionism and cognitive architecture: A critical analysis. Cognition, 28(1-2), 3–71. https://doi.org/10.1016/0010-0277(88)90031-5

Frege, Gottlob. (1948) "Sense and Reference." The Philosophical Review. Durham, NC: Duke University Press.

Friston K. (2009) The free-energy principle: a rough guide to the brain? Trends Cogn Sci. 13(7):293-301. doi: 10.1016/j.tics.2009.04.005.

Gile, D. (1995). Basic Concepts and Models for Interpreting and Translation Training. Philadelphia, PA: John Benjamins Publishing Company.

Gutt, E. A. (2000). Translation and relevance: Cognition and context. Blackwell Publishing: Oxford, UK; St Jerome Publishing: Manchester, UK.

Gutt, E. A. (2005). On the significance of the cognitive core of translation. The Translator, 11(1), 25–49

Halverson, Sandra L. (2017) Gravitational pull in translation. Testing a revised model". Empirical Translation Studies: New Methodological and Theoretical Traditions, edited by Gert De Sutter, Marie-Aude Lefer and Isabelle Delaere, Berlin, Boston: De Gruyter Mouton, 2017, pp. 9-46. https://doi.org/10.1515/9783110459586-002

Halverson, Sandra Louise (2010).Cognitive translation studies: developments in theory and method,Translation and cognition. John Benjamins Publishing Company. ISSN 9789027231918. p. 349–369.

Harnad, Stevan (1990) The symbol grounding problem, Physica D: Nonlinear Phenomena, Volume 42, Issues 1–3, Pages 335-346, https://doi.org/10.1016/0167-2789(90)90087-6.

Harris, Zellig S. (1954) Distributional Structure, WORD, 10:2-3, 146-162, https://doi.org/10.1080/00437956.1954.11659520

Hermans, Theo. (1999) Translation inSystems: Descriptive and SystemicApproaches Explained. TranslationTheories Explained Ser. 7. UnitedKingdom: St. Jerome, 1

House, J. (1997). Translation Quality Assessment: A Model Revisited. Tubingen: Gunter Narr

Jakobsen, A. Lykke (2026). Cognitive Processes in Translation. In H. Nesi, & P. Milin (Eds.), *International Encyclopedia of Language and Linguistic* (Vol. 10, pp. 208-212). Elsevier. https://doi.org/10.1016/B978-0-323-95504-1.01302-8

Kiverstein J.D. and Rietveld E. (2018) Reconceiving representation-hungry cognition: an ecological-enactive proposal. Adapt Behav. 26(4):147-163. doi: 10.1177/1059712318772778.

Kiverstein, J.D. and Rietveld, E. (2020). Scaling-up skilled intentionality to linguistic thought. Synthese. doi.org/10.1007/s11229-020-02540-3

Kripke, Saul. (1979) Naming and Necessity. Oxford: Blackwell.

Lakoff, G., & Johnson, M. (1980). Johnson Metaphors We Live by. Chicago: University of Chicago Press.

Lakoff, G., & Johnson, M. (1999). Philosophy in the Flesh: The Embodied Mind and Its Challenge to Western Thought. New York: Basic Books.

Lenci, Alessandro. 2018. “Distributional Models of Word Meaning.” Annual Review of Linguistics 4: 151–71. DOI : 10.1146/annurev-linguistics-030514-125254

Malmkjær, Kirsten. (1993). “Underpinning Translation Theory”. Target 5:2 (1993), pp. 133–148. https://doi.org/10.1075/target.5.2.02mal | Published online: 1 January 1993

Mikolov, Tomas and Kai Chen and Greg Corrado and Jeffrey Dean. (2013) Efficient Estimation of Word Representations in Vector. Space https://arxiv.org/abs/1301.3781

Millikan, R. G. (1984). Language, Thought, and Other Biological Categories: New Foundations for Realism. Cambridge, MA: MIT Press/ A Bradford Book

Mollo, D. Coelho, Millière R (2026) The vector grounding problem. Philosophy and the Mind Sciences 7(1). https://doi.org/10.33735/phimisci.2026.12307, URL https://philosophymindscience.org/index.php/phimisci/article/view/12307

Nida, E. (1964). Toward a science of translation.Leiden: Brill.

Nord, Christiane. (1997) Translation as a Purposeful Activity: Functionalist Approaches Explained. Manchester: St Jerome Press.

Ouail, Kitoun, Niklas Nolte, and Bouchacourt, Diane and Williams, Adina and Rabbat, Mike and Ibrahim, Mark. (2024) The Factorization Curse: Which Tokens You Predict Underlie the Reversal Curse and More. Advances in Neural Information Processing Systems. (37). Curran Associates, Inc. https://proceedings.neurips.cc/paper_files/paper/2024/file/cbcce87f745072c819204529be843d16-Paper-Conference.pdf

Polanyi, Michael (1967) The Tacit Dimension. Anchor Books, 1967

Putnam, Hilary (1975) “The meaning of ‘meaning’”, In Mind, Language and Reality ed. Putnam. Cambridge: Cambridge University Press.

Qorib, Muhammad Reza and Junyi Li and Hwee Tou Ng (2025) Just Go Parallel: Improving the Multilingual Capabilities of Large Language Models. https://arxiv.org/abs/2506.13044

Quine, Willard Van Orman (2013) [1960]. Word and Object (New ed.). Cambridge, MA: MIT Press. doi:10.7551/mitpress/9636.001.0001. ISBN 9780262518314. OCLC 808006883. New edition with a foreword by Patricia Churchland.

Reiß, Katharina and Vermeer, Hans J. (1984) Grundlegung einer allgemeinen Translationstheorie. Tübingen: Niemeyer

Rhee, Joo Yull. (2026) Beyond Hooking Onto the World: Referential Profiles and the Numerical Structure of LLM Grounding. https://arxiv.org/abs/2606.21195

Rietveld, E., & Kiverstein, J. (2014). A rich landscape of affordances. Ecological Psychology, 26, 325–352. doi: 10.1080/10407413.2014.958035

Risku, H. (2005). Translations- und kognitionswissenschaftliche Paradigmen: Der Mensch im Mittelpunkt. In L. N. Zybatow (Ed.), Translationswissenschaft im interdisziplinären Dialog: Innsbrucker Ringvorlesungen zur Translationswissenschaft III (pp. 55-70,). Peter Lang.

Risku, H., and R. Rogl. (2021). “Translation and Situated, Embodied, Distributed, Embedded and Extended Cognition.” In The Routledge Handbook of Translation and Cognition, ed. by F. Alves, and A. L. Jakobsen, 478–499. London: Routledge.

Risku, H., Schlager, D., & Baumann, A. (2026). Beyond individual mastery: Translation expertise as collective, distributed enaction. Translation & Interpreting Studies: The Journal of the American Translation and Interpreting Studies Association.

Robinson, Douglas (2020), "Reframing translational norm theory through 4EA cognition". Cognition & Behavior 3(1), 122–142.

Robinson, Douglas (2023), *Questions for Translation Studies*. Benjamins Translation Library

Ryle, Gilbert. [1949] 2002. The Concept of Mind. Chicago: University of Chicago Press. p. 327.

Schaeffer , Moritz ; Carl, Michael. 2013. Shared Representations and the Translation Process : A Recursive Model. In: Translation and Interpreting Studies, Vol. 8, No. 2, 2013, p. 169–190,

Searle J.R. (1980) Minds, brains, and programs. Behavioral and Brain Sciences. 3(3):417–424. https://doi.org/10.1017/S0140525X00005756

Searle, J. R. (1958). "Proper Names." Mind, 67(266), 166–173

Seleskovitch, Danica and Marianne Lederer. (1984) Interpréter pour traduire. Paris: Didier Erudition, 1984.

Shackell, Cameron. De Vine, Lance. (2022) Quantifying the genericness of trademarks using natural language processing: an introduction with suggested metrics. Artificial Intelligence and Law, 30. DOI: 10.1007/s10506-021-09291-7

Sperber, D., & Wilson, D. (1986). Relevance. Communication and cognition. Oxford: Basil Blackwell. 2nd edition published in 1995.

Toury, Gideon. (1995) Descriptive Translation Studies and beyond. Amsterdam: John Benjamins.

Tymoczko (1999) Translating in a Postcolonial Context: Early Irish Literature in English Translation

Vervaeke, John ; Timothy P. Lillicrap andBlake A. Richards. (2012) Relevance Realization and the Emerging Framework in Cognitive Science, Journal of Logic and Computation, Volume 22, Issue 1 Pages 79–99, https://doi.org/10.1093/logcom/exp067

Vinay, J. P., & Darbelnet, J. (1995). Comparative stylistics of French and English: A methodology for translation. Amsterdam: John Benjamins Publishing.

Wang, Hetong, Pasquale Minervini, and Edoardo Ponti. (2024). Probing the Emergence of Cross-lingual Alignment during LLM Training. In Findings of the Association for Computational Linguistics: ACL 2024, pages 12159–12173, Bangkok, Thailand. Association for Computational Linguistics.

Wheeler, Samuel (2025) Does ChatGPT refer with Names? Design Intention and Derivative Reference in Large Language Models. Articles, Issue #53, https://nonsite.org/does-chatgpt-refer-with-names-design-intention-and-derivative-reference-in-large-language-models/

Wittgenstein, Ludwig. (1958) Philosophical Investigations. Oxford: Basil Blackwell. Publisher Ltd., 1958.